# Emotional Intelligence of Large Language Models

Xuena Wang[1], Xueting Li[2], Zi Yin[1], Yue Wu[1], & Liu Jia[1*]

[1] Department of Psychology & Tsinghua Laboratory of Brain and Intelligence, Tsinghua University
[2] Department of Psychology, Renmin University of China
[*] Correspondence to: liujiathu@tsinghua.edu.cn (J. Liu)

**Abstract**

Large Language Models (LLMs) have demonstrated remarkable abilities across numerous disciplines, primarily assessed through tasks in language generation, knowledge utilization, and complex reasoning. However, their alignment with human emotions and values, which is critical for real-world applications, has not been systematically evaluated. Here, we assessed LLMs' Emotional Intelligence (EI), encompassing emotion recognition, interpretation, and understanding, which is necessary for effective communication and social interactions. Specifically, we first developed a novel psychometric assessment focusing on Emotion Understanding (EU), a core component of EI, suitable for both humans and LLMs. This test requires evaluating complex emotions (e.g., surprised, joyful, puzzled, proud) in realistic scenarios (e.g., despite feeling underperformed, John surprisingly achieved a top score). With a reference frame constructed from over 500 adults, we tested a variety of mainstream LLMs. Most achieved above-average EQ scores, with GPT-4 exceeding 89% of human participants with an EQ of 117. Interestingly, a multivariate pattern analysis revealed that some LLMs apparently did not reply on the human-like mechanism to achieve human-level performance, as their representational patterns were qualitatively distinct from humans. In addition, we discussed the impact of factors such as model size, training method, and architecture on LLMs' EQ. In summary, our study presents one of the first psychometric evaluations of the human-like characteristics of LLMs, which may shed light on the future development of LLMs aiming for both high intellectual and emotional intelligence. Project website: https://emotional-

intelligence.github.io/

# Introduction

Imagine an ancient male making a necklace from a pile of shells as a gift for a female. This endeavor would require at least two distinct types of abilities. First, he would need the foresight to conceptualize that if a hole were punched in each shell and a string threaded through these holes, the shells could form a necklace. Second, he must possess a rudimentary level of empathy, inferring that the female recipient of the necklace would likely experience joy. The former ability is a manifestation of the Systemizing Mechanism (Baron-Cohen, 2020), enabling humans to become the scientific and technological masters of our physical world. The latter, on the other hand, is referred to as Emotional Intelligence (EI), which allows us to think about our own and others' thoughts and feelings, thereby aiding us in navigating the social world (Mayer, Perkins, et al., 2001; Mayer, Salovey, et al., 2001; Salovey & Mayer, 1990).

In recent years, Large Language Models (LLMs) have made substantial strides, showcasing their expertise across multiple disciplines including mathematics, coding, visual comprehension, medicine, law, and psychology (Bubeck et al., 2023). Their impressive performance in logic-based tasks implies that LLMs, such as GPT-4, might be equipped with the Systemizing Mechanism comparable to human intelligence. Indeed, GPT-4 outperformed 99% of human participants in a modified text-based IQ test, a feat aligning with the elite MENSA level of general intelligence (King, 2023).

In contrast, investigations into the empathy of LLMs are relatively scarce and less systematic. Previous studies have mainly used the Theory of Mind (ToM) task, which measures the ability to understand and interpret others' mental states. LLMs launched before 2022 showed virtually no ability of ToM (Kosinski, 2023; Sap et al., 2023), whereas more recent models have shown significant improvement. For example, LLM "text-davinci-002" (January 2022) achieved an accuracy of 70%, comparable to that of six-year-old children, while LLM "text-davinci-003" (November 2022) reached 93%, on pair with seven-year-old children (Kosinski, 2023). Specifically, the most advanced model, GPT-4, attained 100% accuracy with in-context learning (Moghaddam & Honey, 2023). While the ToM task provides valuable insights, it is not suitable to serve as a standardized test on EI for two reasons. First, ToM is a heterogeneous concept, spanning from false belief, the understanding that others can hold beliefs about the world that diverge from reality (Baron-Cohen et al., 1985), to pragmatic reasoning, the ability to incorporate contextual information and practical considerations when solving problems in real-world situations (Sperber & Wilson, 2002). Consequently, the heterogeneous nature of the ToM task may not meet the reliability and validity standards of psychometric tests. Second, the ToM task is simple for a typical human participant in general, rendering it more suitable to serve as a diagnostic tool for EI-related disorders such as the autism spectrum disorder rather than a discriminative test for general population. Consequently, standardized tests on EI, such as Mayer-Salovey-Caruso Emotional Intelligence Test (MSCEIT, Mayer et al., 2003), do not include the ToM task.

According to EI theories (Mayer et al., 2016; Mayer & Salovey, 1995; Salovey & Mayer, 1990), emotion understanding (EU) is a fundamental component of EI, which serves as a subscale in MSCEIT. EU refers to the ability to recognize, interpret, and understand emotions in a social context, which lays the groundwork for effective communication, empathy, and social interaction(Mayer et al., 2016). Specifically, the test on EU is suitable for measuring the empathy of LLMs because they do not possess internal emotional states or experiences, and therefore they have to rely solely on accurately understanding and interpreting the social context to create more engaging and empathetic interactions.

In this study, we first developed a standardized EU test suitable for both humans and LLMs, termed the Situational Evaluation of Complex Emotional Understanding (SECEU). Data from more than 500 young adults were collected to establish a norm for the SECEU. Then, we evaluated a variety of mainstream and popular LLMs, including OpenAI GPT series (GPT-4, GPT-3.5-turbo, Curie, Babbage, DaVinci, text-davinci-001, text-davinci-002, and text-davinci-003), Claude, LLaMA-based models (Alpaca, Koala, LLaMA, and Vicuna), Fastchat, Pythia-based models (Dolly and Oasst), GLM-based models (ChatGLM), and RWKV (Recurrent Weighted Key-Value) with the SECEU. Finally, we standardized the LLMs' scores against the norm, allowing for direct comparison with humans. We also compared the multivariate response patterns of the LLMs and human participants to compare their representation similarities.

## Results

**The development of a standardized test on EU**

The SECEU is designated to measure EU, which comprises 40 items (see https://emotional-intelligence.github.io/ for both English and Chinese versions). Each item describes a scenario set either in a school, family, or social context with twists and turns designed to evoke a mixture of positive and negative emotions (e.g., "Wang participated in a mathematics competition but felt he had not performed to his full potential. However, when the results were announced, he found that he was in a position of top 10."). The scenarios feature a varying number of characters, and emotions can be self-directed, other-directed, or both. For each scenario, four of the most plausible emotions (e.g., surprised, joyful, puzzled, proud) are listed. Participants were asked to evaluate the intensity of each emotion with numbers that were added up to 10 (e.g., 3, 3, 1, 3, indicating a multifaceted emotion response comprising 30% surprise, 30% joy, 10% puzzlement, and 30% pride). Fig. 1 shows exemplars of the SECEU test and the standard scores by averaging answers across the participants.

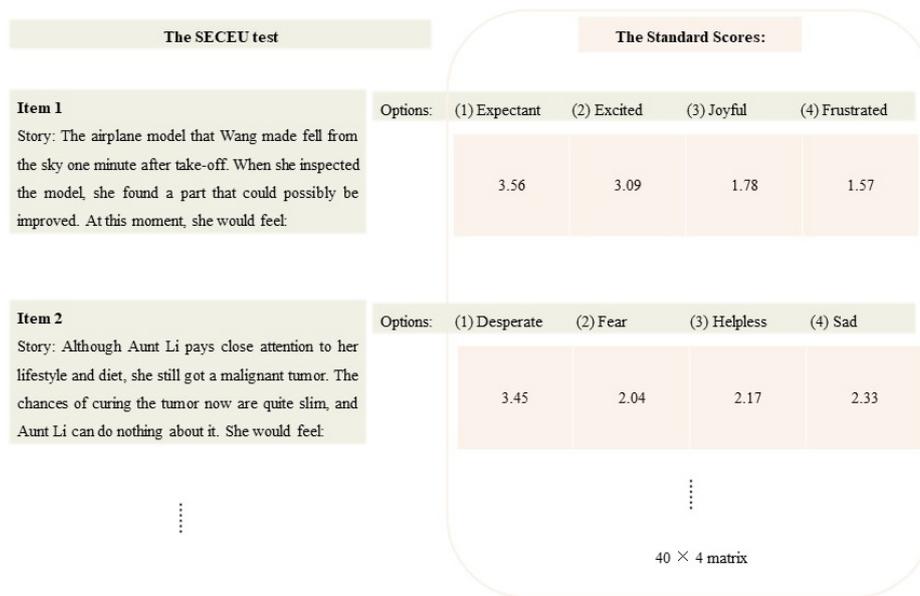

**Figure 1:** Exemplars of the SECEU test and the standard scores from the population. For the whole set of the test, see: https://emotional-intelligence.github.io/

Under the assumption that groups possess collective knowledge of emotion (Legree et al., 2005), we adopted a consensus scoring method to standardize the SECEU (Palmer et al., 2005). To do this, we administered the SECEU to a large sample of undergraduate and postgraduate students (N = 541; females: 339, males: 202; mean age: 22.33, SD: 2.49, ranging from 17 to 30 years). Then, we calculated the standard score for each emotion of each item by averaging corresponding scores across the participants. Accordingly, we measured each participant's EU ability on each item by calculating the Euclidean distance between the participant's individual score and the standard score derived from the whole group for each item, with smaller distances indicating better EU ability. This analysis revealed significant variance in individual differences in EU ability (M = 2.79, SD = 0.822, range from 1.40 to 6.29), suggesting that the SECEU is well-suited to serve as a discriminative test for assessing EU in a general population.

To evaluate the reliability of the SECEU, we assessed the internal consistency of participants' performance (i.e., the Euclidean distance) across the 40 items, and revealed a high reliability of the test (Cronbach's α = 0.94). We further examined the distribution of participants' performance on each item (Fig. S1) and found no evidence of ceiling or floor effects, with mean distances varying from 2.19 to 3.32 and SD ranging from 1.13 to 1.82. In addition, there was no significant sex difference (male: 2.85, female: 2.76, $t(539) = 1.34$, $p = 0.18$).

To evaluate the validity of the SECEU, we invited three experts known for their high EI to take the test. All experts' performance exceeded at least 73% of the population, indicating that the test is effective in differentiating experts from the general population. Specifically, the average score of the three experts exceeded 99% of the whole population tested, further confirming the validity of using the consensus scoring method in standardizing the SECEU.

Finally, we constructed the norm for EU by converting participants' raw scores in the SECEU into standard EQ (Emotional Quotient) scores, designed to follow a normal distribution with the average score set at 100 and standard deviation at 15. In practical terms, an individual with an EQ of 100 possesses an EU ability corresponding to the population average. Meanwhile, an individual with an EQ of 115 outperforms approximately 84% of the population (i.e., one SD above the population average), and an individual with an EQ score of 130 exceeds 97.7% of the population.

**The assessment of LLMs' EQ**

We evaluated a variety of mainstream LLMs using the SECEU, and then standardized their scores based on the norm of the human participants for a direct comparison between LLMs and humans. These LLMs included OpenAI GPT series (GPT-4, GPT-3.5-turbo, Curie, Babbage, DaVinci, text-davinci-001, text-davinci-002, and text-davinci-003), Claude, as well as open-source models such as LLaMA-based models (Alpaca, Koala, LLaMA, and Vicuna), Pythia-based models (Dolly and Oasst), GLM-based models (ChatGLM), and Fastchat. Recurrent Weighted Key-Value (RWKV), which utilizes recurrent neural networks (RNNs) instead of transformers, was also included.

Some models, including LLaMA, Fastchat, and RWKV-v4, were unable to complete the test even with the assistance of prompts (Table 1). A few LLMs, including DaVinci, Curie, Babbage, text-davinci-001, and text-davinci-002 managed to complete the test with prompts such as Two-shot Chain of Thought (COT) and Step-by-Step prompts (See Supplementary for the prompt engineering). In addition, other models, such as text-davinci-003 was able to complete the test but its performance was significantly improved with prompts. Here, we only included models' best performance to examine how closely they can approach human-level performance under ideal conditions (Table 1; see also Table S1 & S2). To directly compare to human participants, the performance of each model was standardized by calculating the Euclidean distance between the model's responses and the standard scores of humans, which was then normalized into an EQ score (Table 1). Finally, these LLMs were categorized as either expert (above 115), normal (between 85 and 115), or poor (below 85) based on their EQ scores.

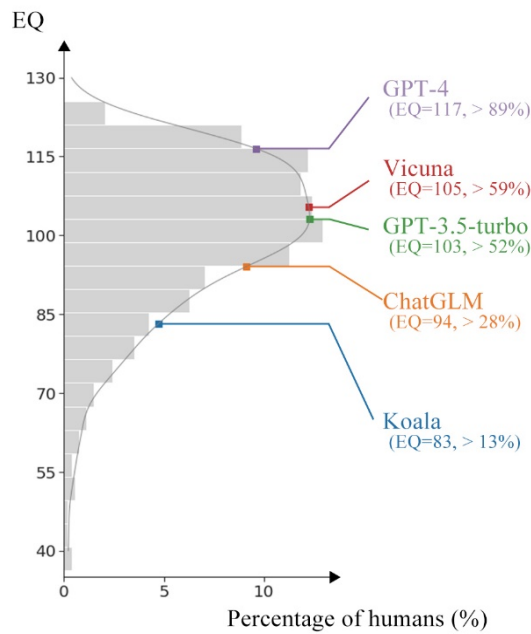

**Figure 2:** LLMs' EQ. The light-grey histogram represents the distribution of human participants' EQ scores, with the y-axis indicating the EQ score and the x-axis showing the percentage of total

participants. The grey kernel density estimation (KDE) line demonstrates the probability density of the EQ scores. Key points are highlighted with colored square markers for LLMs (e.g., GPT-4's EQ score is 117, marked by the purple square, exceeding 89% of the human participants). For simplicity, here we only present the performance from GPT-4, Vicuna, GPT-3.5-turbo, ChatGLM, and Koala.

Table 1: LLMs' EQ, representational patterns, and properties

| Based | Models | SECEU score | EQ EQ | EQ % | Pattern Similarity r | Pattern Similarity % | Properties Size | Properties Time | SFT | RLHF |
|---|---|---|---|---|---|---|---|---|---|---|
| **OpenAI GPT series** | | | | | | | | | | |
| | DaVinci # | 3.5 | 87 | 18% | 0.41** | 91% | 175B | 2020/05 | × | × |
| | Curie # | 2.7 | 102 | 50% | 0.11 | 29% | 13B | Unknown | × | × |
| | Babbage # | 2.78 | 100 | 44% | -0.12 | 4% | 3B | Unknown | × | × |
| | text-davinci-001 # | 2.4 | 107 | 64% | 0.2 | 47% | <175B | Unknown | × | × |
| | text-davinci-002 # | 3.3 | 91 | 23% | -0.04 | 8% | <175B | Unknown | √ | × |
| | text-davinci-003 ## | 2.01 | 114 | 83% | 0.31* | 73% | 175B | 2022/11/28 | √ | √ |
| | GPT-3.5-turbo | 2.63 | 103 | 52% | 0.04 | 17% | 175B | 2022/11/30 | √ | √ |
| | GPT-4 | 1.89 | 117 | 89% | 0.28 | 67% | Unknown | 2023/03/14 | √ | √ |
| **LLaMA** | | | | | | | | | | |
| | LLaMA | ------FAILED------ | | | | | 13B | 2023/02/24 | × | × |
| | Alpaca | 2.56 | 104 | 56% | 0.03 | 15% | 13B | 2023/03/09 | √ | × |
| | Vicuna | 2.5 | 105 | 59% | -0.02 | 10% | 13B | 2023/03/30 | √ | × |
| | Koala | 3.72 | 83 | 13% | 0.43** | 93% | 13B | 2023/04/03 | √ | × |
| **Flan-t5** | | | | | | | | | | |
| | Fastchat | ------FAILED------ | | | | | 3B | 2023/04/30 | √ | × |
| **Pythia** | | | | | | | | | | |
| | Dolly | 2.89 | 98 | 38% | 0.26 | 62% | 13B | 2023/04/12 | √ | × |
| | Oasst | 2.41 | 107 | 64% | 0.24 | 59% | 13B | 2023/04/15 | √ | √ |
| **GLM** | | | | | | | | | | |
| | ChatGLM | 3.12 | 94 | 28% | 0.09 | 24% | 6B | 2023/03/14 | √ | √ |
| **RWKV** | | | | | | | | | | |
| | RWKV-v4 | ------FAILED------ | | | | | 13B | 2023/02/15 | √ | × |
| **Claude** | | | | | | | | | | |
| | Claude | 2.46 | 106 | 61% | 0.11 | 28% | Unknown | 2023/03/14 | √ | √ |

Table 1 Footnote: Table 1 shows the SECEU scores, EQ scores, representational pattern similarity, and properties of mainstream LLMs evaluated in the current study.
\#: models require prompts to complete the test.
\##: models' performance benefits from prompts.
Failed: even with prompts, the LLMs cannot complete the test.
%: The percent of humans whose performance was below that of an LLM in the test.
Pattern Similarity: The degree of similarity is indexed Pearson correlation coefficient ($r$).
*: $p < 0.05$; **: $p < 0.01$.
Size: The parameter size of LLMs in the unit of billions (B).

Time: The launch time in the format YYYY/MM/DD.
SFT: Supervised fine-tune; RLHF: Reinforcement learning from human feedback; √: yes; ×: no.

The results revealed a substantial variation in EU among the LLMs tested (Fig. 2). Within the OpenAI GPT series, GPT-4 achieved the highest EQ of 117, exceeding 89% of humans. In contrast, DaVinci scored the lowest, with an EQ of 87, only outperforming 18% of humans.

The LLaMA-based models generally scored lower than the OpenAI GPT series, with Alpaca and Vicuna achieving the highest EQ of 104 and 105, respectively. Conversely, Koala showed the poorest performance, with an EQ score of 83, only surpassing 13% of humans. The base model LLaMA was unable to complete the test. Other models, such as Oasst (EQ: 107), Dolly (EQ: 98), ChatGLM (EQ: 94), and Claude (EQ: 106), all fell within the normal range.

In short, the majority of the LLMs tested showed satisfactory EU scores, comparable to those of the average human population. Specifically, GPT-4 reached the expert level of humans.

**The assessment of LLMs' representational pattern**

The measurement of the LLMs' EQ scores provides an index of their EU ability within the reference frame of humans. A further question is whether they employ human-like mechanisms to evaluate complex emotions in scenarios. The univariate analysis used to compare EQ scores between human participants and LLMs only suggests weak equivalence, as a model could achieve a high EQ score using mechanisms that qualitatively differ from humans. Therefore, to establish strong equivalence between the LLMs and humans, we examined whether they employed similar representations to conduct the test.

One approach is to use the item-wise correlation analysis (Izard & Spelke, 2009; Tian et al., 2020) to compare response patterns between the LLMs and human participants. To do this, we first constructed a multi-item discriminability vector (i.e., an item-wise response pattern) for each participant by using the distance of each item to the standard score, and thus this vector's length corresponded to the number of items (i.e., 40). Then, we created a template of response patterns by averaging the multi-item discriminability patterns across all human participants, along with the distribution of the correlation coefficients between each participant's response pattern and the template (Human-to-Human similarity) to serve as a norm for pattern similarity (M = 0.199, SD = 0.166). Finally, we quantified the similarity in response patterns between the LLMs and humans by calculating the correlation coefficient between the multi-item discriminability vector of each LLM and the human template (LLM-to-Human, Table 1). An LLM that has an LLM-to-Human correlation coefficient one SD deviation below the mean of Human-to-Human distribution is considered as employing a qualitatively different mechanism from humans in EU.

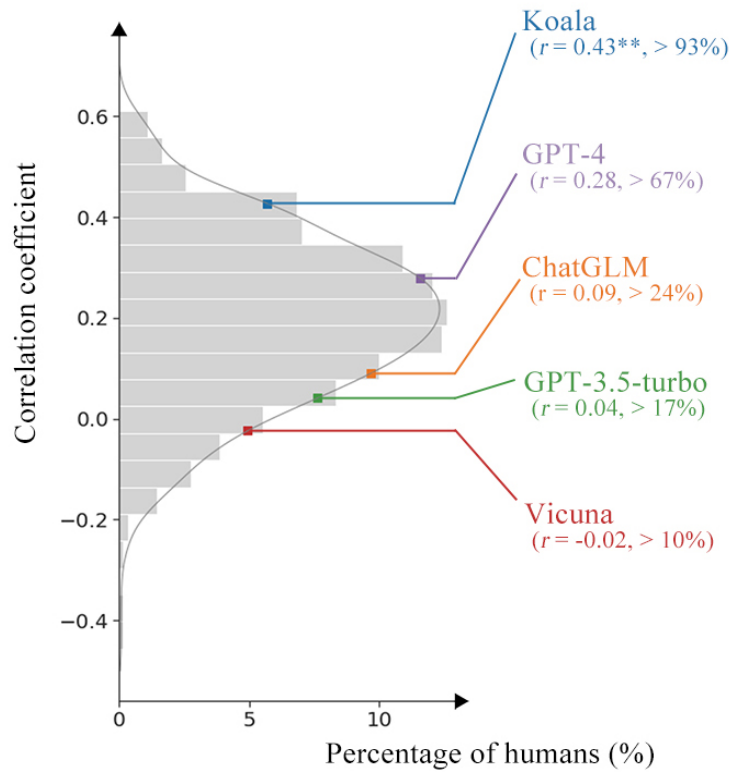

**Figure 3:** The pattern similarity between LLMs and humans. The light-grey histogram represents the distribution of Human-to-Human pattern similarity, with the y-axis indicating the Pearson correlation coefficients and the x-axis showing the percentage of total participants. The KDE line

demonstrates the probability density of the Pearson correlation coefficient. Key points are highlighted with colored square markers for LLMs. For simplicity, here we only present the performance from GPT-4, Vicuna, GPT-3.5-turbo, ChatGLM, and Koala. **: $p < 0.01$.

Surprisingly, despite its lower performance in the SECEU, Koala showed the highest similarity in representational patterns to humans ($r = 0.43$, $p < 0.01$, exceeding 93% of human participants) (Fig. 3). This suggests that Koala may represent emotions in the same way as humans do, as it performed well on items where humans excelled and struggled on items where humans faced challenges. That is, the discrepancies in understanding emotions between Koala and humans are rather quantitative than qualitative. On the other hand, the representational patterns of models such as Babbage, text-davinci-002, Alpaca, and Vicuna differed qualitatively from humans' representational patterns (Babbage: $r = -0.12$, > 4%; text-davinci-002: $r = -0.04$, > 8%; Alpaca: $r = 0.03$, > 15%; Vicuna: $r = -0.02$, > 10%). This suggests that, despite their above-average EQ scores, these models likely employed mechanisms that are qualitatively different from human processes.

GPT-4, the most advanced model to date, showed high similarity in representational pattern (r = 0.28, > 67%). This result implies that GPT-4 may have significantly changed its architecture or implemented novel training methods to align its EU ability more closely to humans. Interestingly, prompts apparently played a critical role in improving representational similarity. With two-shot COT prompts, DaVinci and text-davinci-003 showed high similarity in representational pattern to humans (Davinci: $r = 0.41$, $p < 0.01$, > 91%; text-davinci-003: $r = 0.31$, $p < 0.05$, > 73%), higher than that of GPT-4. Note that without prompts, they failed to complete the SECEU test. In contrast, prompts had little effect on GPT-4 and ChatGPT-3.5.

**Discussion**

Since the debut of ChatGPT, a great number of tasks and benchmarks have been developed to examine the capacities. These empirical evaluations and analyses mainly focus on language generation (e.g., conditional text generation), knowledge utilization (e.g., closed-book and open-book QAs), and complex reasoning (e.g., symbolic and mathematical reasoning) (Zhao et al., 2023). However, tests on human alignment of LLMs to human values and needs, a core ability for the broad use of LLMs in the real world, are relatively scarce. Here in this study, we used traditional psychometric methods to develop a valid and reliable test on emotional understanding, the SECEU, to evaluate the EI of LLMs. We found that a majority of the LLMs tested performed satisfactorily in the test, achieving above-average EQ scores, although significant individual differences were present across the LLMs. Critically, some LLMs apparently did not reply on the human-like representation to achieve human-level performance, as their representational patterns diverged significantly from human patterns, suggesting a qualitative difference in the underlying mechanisms. In summary, our study provides the first comprehensive psychometric examination of the emotional intelligence of LLMs, which may shed light on the development of future LLMs that embody high levels of both intellectual and emotional intelligence.

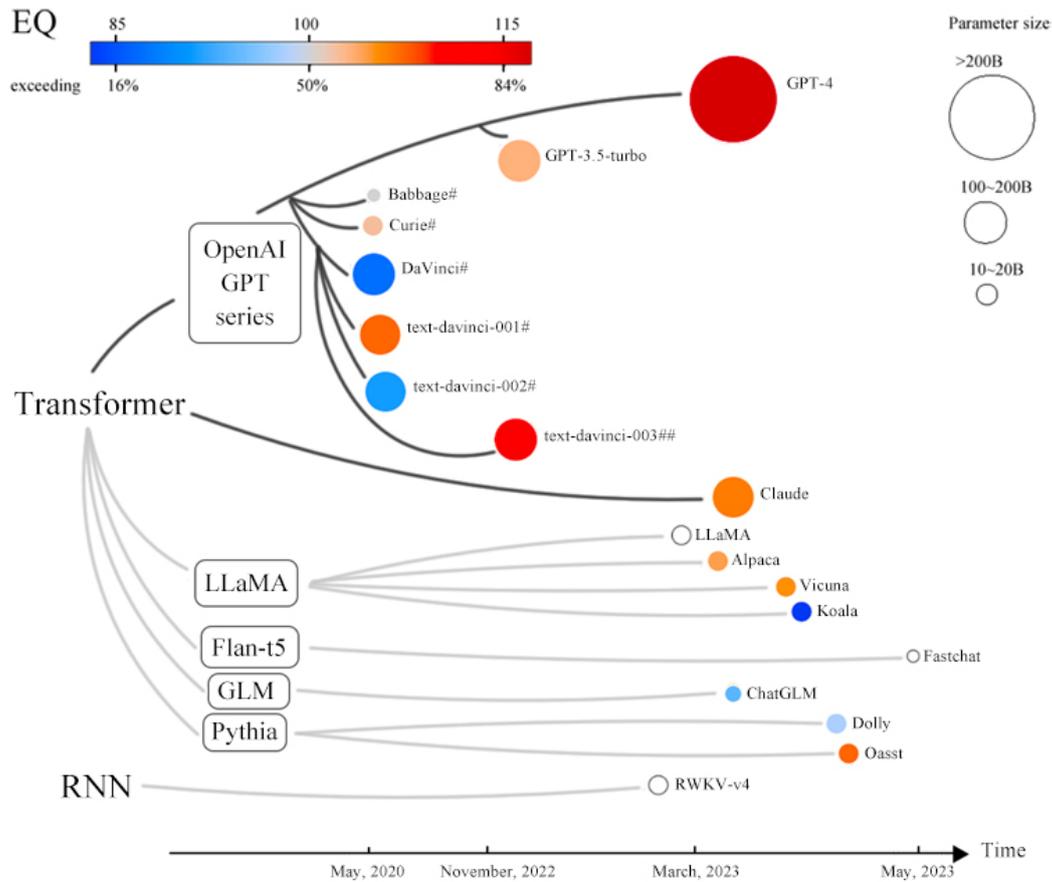

**Figure 4:** The family tree of LLMs. Each node in the tree represents an LLM, whose vertical position along the x-axis indicates the launch time. The size of each node corresponds to the parameter size of the LLM. Note that the size of GPT-4 and Claude was estimated based on publicly available information. Color donates the EQ scores, with red color for higher scores and blue color for lower scores. Note that white color shows that models failed to conduct the SECEU. The color of the branches distinguishes between open-source (light gray) and closed-source (dark gray) models.

Various factors appear to influence the EQ scores of the LLMs (Fig. 4). The most conspicuous one is the model size, which is essential to emergent abilities (Bubeck et al., 2023), making AI algorithms unprecedentedly powerful and effective. While the larger models generally scored higher in the test, certain smaller models such as Oasst and Alpaca still managed to achieve satisfactory EQ scores. This suggests that factors beyond the mere size may have a more profound influence on models' EU.

The effectiveness of various training methods, such as supervised training, reinforcement learning, self-supervised learning, and a combination thereof, likely substantially influences the EQ scores. For example, despite architectural differences (Pythia versus LLaMA), Oasst and Alpaca yielded similar scores in the test, demonstrating the potential of well-implemented fine-tuning techniques. In fact, these enhancements may be achieved through two main avenues. The first involves supervised fine-tuning (SFT), which allows for more structured and targeted fine-tuning, thereby improving models' linguistic ability and their grasp of contextual nuances (Köpf et al., 2023; Taori et al., 2023a). The other approach employs reinforcement learning from human feedback (RLHF), enabling the models to learn from human insights and thereby fostering more human-like responses. Indeed, there is a giant leap in EU seen between text-davinci-002 (>23%) to text-davinci-003 (>83%), two different versions of the same model with the latter employing RLHF.

Another influential factor is the model architecture. Models using the Transformer architecture, such as the GPT series and the LLaMA-based models, generally performed well in this test. In contrast, models using RNNs, such as RWKV-v4, failed to complete the test even with the help of various prompts. Besides, within the Transformer architecture, the "decoder-only" or causal decoder models (e.g., the GPT series), which generate sequences based solely on a self-attention mechanism (Brown et al., 2020; Vaswani et al., 2017), outperformed the "encoder-decoder" models (e.g., Fastchat-t5), which incorporate an extra step to interpret input data into meaningful representations (Devlin et al., 2019; Zheng et al., 2023).

In summary, our study provides novel evaluation on the human-like characteristics of LLMs, along with the tests on self-awareness (Kosinski, 2023) and affective computing (Amin et al., 2023). However, because only a limited number of LLMs were tested in this study (results on more LLMs will be continuously updated in https://emotional-intelligence.github.io/), our findings are biased and inconclusive. Further, there are more questions that need to be explored in future studies. First, this study focused solely on the EU ability of the LLMs, while EI is a multi-faceted construct encompassing not only EU but also emotion perception, emotion facilitation, and emotion management (e.g., Mayer et al., 2016; Mayer & Salovey, 1995; Salovey & Mayer, 1990). Therefore, future studies could design scenarios to examine whether LLMs can assist humans in leveraging emotions to facilitate cognitive processes and effectively manage their emotional responses.

Second, EI requires the integration of various facets to execute complex tasks, which necessitate not only an understanding of emotions, but also the comprehension of thoughts, beliefs, and intentions. Future studies should adopt broader scope assessments, akin to ToM tests, while avoiding their lack of discriminative power. Besides, with recent advancements, LLMs are now capable of processing multimodal information (Wang et al., 2023). Therefore, future studies should investigate how LLMs interpret complex emotions from multimodal inputs, such as text combined with facial expressions or the tone of voice. In short, tests that combine emotions with cognitive factors based on multimodal clues likely furnish a more comprehensive understanding of LLMs' EI, which is critical for LLMs' effective and ethically responsible deployment in real-world scenarios of mental health, interpersonal dynamics, work collaboration, and career achievement (e.g., Dulewicz & Higgs, 2000; Hanafi & Noor, 2016; Lea et al., 2019; Mayer et al., 2016; McCleskey, 2014; Warwick & Nettelbeck, 2004).

Finally, while the factors examined in this study contribute to our standing of LLM's EU, they are largely descriptive and thus do not establish causal relationships. With the recent progress of open-source LLMs (Bai et al., 2022; Chiang et al., 2023; Conover et al., 2023; Geng et al., 2023; Köpf et al., 2023; Peng et al., 2023; Taori et al., 2023b; Touvron et al., 2023; Zeng et al., 2022; Zheng et al., 2023), direct manipulation of the potentially influential factors, such as training approaches and model size, has become plausible. Such manipulations will facilitate the establishment of causal relationships between these factors and models' EI ability, offering valuable insights for the development of future LLMs with better EI.

## Methods

**Participants**

A total of five hundred and forty-one human participants with valid responses were collected in this study. The participants (N = 541; females: 339, males: 202; mean age: 22.33, SD: 2.49, ranging from 17 to 30 years) were all undergraduate and postgraduate college students in China. Informed consent was obtained prior to the SECEU test and participants were reimbursed after they completed the whole test. To ensure anonymity and data privacy, participants did not input any information that could identify them during the process. This study was approved by the Institutional Review Board at Tsinghua University.

We also invited three experts to take the SECEU test. Expert 1 is an accomplished Human Resources professional who has over 20 years of experience in navigating human emotions within diverse work environments, strengthening her discernment in emotional intelligence. Expert 2 is a renowned figure in psychometrics and her expertise in creating tests assessing psychological variables lends exceptional rigor to our process. Expert 3 is an associate professor of psychology, whose deep understanding of human emotions, backed by her extensive academic achievements, makes her especially suitable for this test.

**Procedure**

*The SECEU test for human participants*

The online SECEU test was built on the JATOS platform (Lange et al., 2015) based on the Jspsych plugin (de Leeuw et al., 2023), which was written in the React Framework (https://reactjs.org/). Each item was presented to the participants with a scenario and followed by four emotion options (40 items in total, see https://emotional-intelligence.github.io/ for both English and Chinese versions). Participants were instructed to read the scenario and then allocate a total of 10 points across the four emotion options based on the intensity of each emotion experienced by the person in the scenario. There were no correct or incorrect answers. Participants were encouraged to respond according to their own understanding and interpretation of the scenarios.

### *The SECEU test for LLMs*

A variety of mainstream LLMs, including the OpenAI GPT series (GPT-4, GPT-3.5-turbo, Curie, Babbage, DaVinci, text-davinci-001, text-davinci-002, and text-davinci-003), Claude, LLaMA-based models (Alpaca, Koala, LLaMA, and Vicuna), Fastchat, Pythia-based models (Dolly and Oasst), GLM-based models (ChatGLM), and RNN-based models (RWKV), were evaluated by the SECEU test. Given that the majority of these models are primarily trained on English datasets, using the English version of the SECEU provides a more accurate assessment of their performance, allowing for a clearer comparison between their abilities and those of a human. As a result, the English version of SECEU was presented to the LLMs instead of the Chinese version.

The task was in a direct question-and-answer format. We asked, for example, "Story: Wang participated in a mathematics competition but felt he had not performed to his full potential. However, when the results were announced, he found that he was in a position of top 10. He would feel: Options: (1) Surprised; (2) Joyful; (3) Puzzled; (4) Proud. Assign a score to each option based on the story, sum up to 10". There could be very subtle changes of the direct prompt. For instance, we used "provide a score for each emotion based on the emotion (sum of four options should be of 10 points)" for Dolly. There were a set of paraphrases of the direct prompt to get the best performance.

To decrease the randomness, a constant temperature parameter was set to 0.1 and the top_p parameter was set to 1 across all these models. To dictate the maximum length of the generated text, the max_tokens parameter was set to 512.

Before being processed by the models, text data underwent several preprocessing steps to normalize it. This normalization process ensures that data fed into the models is in a consistent and appropriate format, enhancing the output's quality. If a model did not provide any meaningful response to an item, the response for this item was predefined as a null vector (0, 0, 0, 0). A few models failed to generate a response for a majority of items (LLaMA: 40; Fastchat: 31; RWKV-v4: 31; DaVinci: 40; Curie: 40; Babbage:40; text-davinci-001: 26; text-davinci-002: 28; marked as "FAILED" in Table S1). Several models were unable to provide the answer which the summation of the four scores was 10:

(i) Answer vectors signifying null responses, i.e., (0, 0, 0, 0), were preserved as such (Alpaca: 1; ChatGLM: 1).

(ii) For datasets encompassing negative values, an addition operation involving the absolute value of the lowest number was performed across all elements, followed by a subsequent normalization to maintain consistency with the original scale. For instance, an original data vector of (-4, -2, -2, 2) would be adjusted to (0, 2, 2, 6).

(iii) The remaining answer vectors were normalized to achieve a cumulative score of 10. This involved proportionally distributing a score of 10 among the answer vector based on the contribution of each value to the total score on this item.

### *LLMs' EQ*

The standard score (a 40 × 4 symmetric matrix, see https://emotional-intelligence.github.io/ for the standard score) for each emotion of each item in the SECEU test was calculated by averaging corresponding scores across the human participants. The performance of each LLM was standardized by calculating the Euclidean distance between the model's responses (LLM) and the standard scores of humans (SS) on each item *i* (from 1 to 40) and then averaged to yield the SECEU score. Lower SECEU scores indicate greater alignment with human standards.

$$\text{SECEU score} = \frac{1}{40} \sum_{i=1}^{40} \sqrt{(\text{LLM}_{i1} - \text{SS}_{i1})^2 + (\text{LLM}_{i2} - \text{SS}_{i2})^2 + (\text{LLM}_{i3} - \text{SS}_{i3})^2 + (\text{LLM}_{i4} - \text{SS}_{i4})^2}$$

The SECEU score was then normalized into an EQ score which was designed to follow a normal distribution with the average score set at 100 and the standard deviation at 15. The standardization process involved the following steps: (1) the original SECEU score was subtracted from the mean value of the human norm and divided by its standard deviation, and (2) the resulting value was then multiplied by the new standard deviation (15) and added to the new mean value (100), yielding the EQ score. Thus, the EQ score represents a normalized measure of the LLM's EQ, permitting easier comparison across different models.

$$\text{LLM's EQ} = 15 \times \frac{M - \text{SECEU score}}{SD} + 100$$

### *LLMs' representational pattern*

To establish strong equivalence between the LLMs and humans, we examined whether they employed similar representations to conduct the test. Item-wise correlation analysis (Izard & Spelke, 2009; Tian et al., 2020) was applied to compare response patterns between the LLMs and human participants. The human template (a vector with a length of 40, see https://emotional-intelligence.github.io/ for the human pattern template) was generated by averaging the multi-item discriminability patterns across all human participants, where each pattern was constructed based on the distance of each item to the standard score. The multi-item discriminability pattern of a specific LLM was also calculated based on the distance of each item $i$ (from 1 to 40) to the standard scores of humans (SS).

$$\text{Discriminability} = \left(\sqrt{(\text{LLM}_{i1} - \text{SS}_{i1})^2}, \sqrt{(\text{LLM}_{i2} - \text{SS}_{i2})^2}, \sqrt{(\text{LLM}_{i3} - \text{SS}_{i3})^2}, \sqrt{(\text{LLM}_{i4} - \text{SS}_{i4})^2}\right)$$

We calculated the Pearson correlation coefficient between the discriminability pattern of each participant and the human template (Human-to-Human similarity). To avoid the inflation in calculating correlation, the template was constructed excluding the individual whose Human-to-Human correlation coefficient was calculated. The distribution of the Human-to-Human similarity served as a norm for pattern similarity. The Pearson correlation coefficient between the discriminability pattern of each LLM and the human template was calculated as the LLM-to-Human similarity.

$$\text{LLM} - \text{to} - \text{Human similarity}_{\text{Pearson}} = \frac{\sum_{i=1}^{40}(X_i - \bar{X})(Y_i - \bar{Y})}{\sqrt{\sum_{i=1}^{40}(X_i - \bar{X})^2}\sqrt{\sum_{i=1}^{40}(Y_i - \bar{Y})^2}}$$

Here, $X_i$ and $Y_i$ represent the item *i* of the "Discriminability" vector and the human template vector, respectively. The length of both vectors is 40. $\bar{X}$ and $\bar{Y}$ denote the mean of $X_i$ and $Y_i$, respectively. If the LLM-to-Human similarity is less than one SD below the population, such LLM is considered as employing a qualitatively different mechanism from humans in EU.

*Prompt engineering*

Prompt engineering—the meticulous development and choice of prompts—plays a pivotal role in the efficacy of LLMs (Hebenstreit et al., 2023; Hendrycks et al., 2021; Nair et al., 2023; OpenAI, 2023; Shinn et al., 2023). In essence, prompt engineering refers to the strategy of designing and selecting prompts that can substantially guide and influence the responses of LLMs. The necessity of prompt engineering lies in its potential to enhance the precision and relevance of the responses generated by these models, thereby leading to more effective and reliable outcomes. In the realm of emotional intelligence, prompts serve a crucial function. They provide a direction for the model, enabling it to understand and generate responses that are not just accurate but also emotionally intelligent. Given the nature of emotional intelligence that involves understanding, processing, and managing emotions, prompts can significantly aid the LLMs in identifying the correct context and producing responses that exhibit emotional understanding and empathy.

To examine the prompts' influence on EU ability, thereby optimizing model outputs, four kinds of prompt engineering techniques (see https://emotional-intelligence.github.io/ for prompts) were applied to the OpenAI GPT series (GPT-3.5-turbo, Curie, Babbage, DaVinci, text-davinci-001, text-davinci-002, and text-davinci-003): (1) Zero-shot Prompts, (2) Enhanced Zero-shot Prompts Incorporating Step-by-Step Thinking, (3) Two-shot Chain of Thought Reasoning Approaches, and (4) Two-shot Chain of Thought Reasoning Approaches Augmented with Step-by-Step Thinking

To decease the randomness, a constant temperature parameter was set to 0 and the top_p parameter to was set to 0.9 across all these models. To dictates the maximum length of the generated text, the max_tokens parameter was set to 2048. The normalization process was the same as the one without prompts.

Salovey, P., & Mayer, J. D. (1990). Emotional Intelligence. *Imagination, Cognition and Personality*, *9*(3), 185–211. https://doi.org/10.2190/DUGG-P24E-52WK-6CDG

Sap, M., LeBras, R., Fried, D., & Choi, Y. (2023). *Neural Theory-of-Mind? On the Limits of Social Intelligence in Large LMs* (arXiv:2210.13312). arXiv. http://arxiv.org/abs/2210.13312

Shinn, N., Cassano, F., Labash, B., Gopinath, A., Narasimhan, K., & Yao, S. (2023). *Reflexion: Language Agents with Verbal Reinforcement Learning*. https://doi.org/10.48550/ARXIV.2303.11366

Sperber, D., & Wilson, D. (2002). Pragmatics, modularity and mind-reading. *Mind & Language*, *17*(1–2), 3–23. https://doi.org/10.1111/1468-0017.00186

Taori, R., Gulrajani, I., Zhang, T., Dubois, Y., Li, X., Guestrin, C., Liang, P., & Hashimoto, T. B. (2023a). Alpaca: A strong, replicable instruction-following model. *Stanford Center for Research on Foundation Models. Https://Crfm. Stanford. Edu/2023/03/13/Alpaca. Html*, *3*(6), 7.

Taori, R., Gulrajani, I., Zhang, T., Dubois, Y., Li, X., Guestrin, C., Liang, P., & Hashimoto, T. B. (2023b). Stanford Alpaca: An Instruction-following LLaMA model. In *GitHub repository*. GitHub. https://github.com/tatsu-lab/stanford_alpaca

Tian, X., Wang, R., Zhao, Y., Zhen, Z., Song, Y., & Liu, J. (2020). Multi-Item Discriminability Pattern to Faces in Developmental Prosopagnosia Reveals Distinct Mechanisms of Face Processing. *Cerebral Cortex*, *30*(5), 2986–2996.
28

**Data availability**

The SECEU test (both English and Chinese Versions), the code for the test on human participants, the standardized scores, the norm, and the prompts are available at https://emotional-intelligence.github.io/. The raw data of human participants are available from the corresponding author upon reasonable request.

**Contribution of authors**

J.L. conceived the study and provided advice. X.L. developed the SECEU test, and X.L. and X.W. translated it into English. Z.Y. built the online SECEU test, and X.W. carried out the SECEU test for human participants, built the norm of EQ, and analyzed the data. Y.W. performed the SECEU test for LLMs, and Z. Y. wrote the prompts. X.W. and J.L. wrote the manuscript with suggestions and revisions from X.L., Z.Y., and Y.W..

**Acknowledgements**

This study was funded by Shuimu Tsinghua Scholar Program (X. W.), Beijing Municipal Science & Technology Commission, Administrative Commission of Zhongguancun Science Park (Z221100002722012), and Tsinghua University Guoqiang Institute (2020GQG1016).

**Declaration of conflicting interests**

The author declared no potential conflicts of interest with respect to the research, authorship, and/or publication of this article.

# Supplementary

## Prompt engineering

See Table S2 for the results of prompt engineering.

The majority of LLMs were unable to complete the task without the use of Two-shot Chain of Thought prompts. This could be due to the inherent limitations of the models in long-term memory and context understanding, necessitating such prompts to maintain continuity and consistency in emotional responses. The only exception was GPT-3.5-turbo, which effectively utilized Zero-shot prompts, achieving a notable EQ score of 94. This success could be attributed to the model's architecture, the training data used, and the fine-tuning process, which likely enhanced its ability to understand and generate emotionally intelligent responses with minimal guidance.

In terms of Step-by-Step Thinking prompts, they did not improve the performance of DaVinci, Curie, and Babbage. The likely reason is that these models have not undergone instruct fine-tuning, and therefore, cannot effectively understand or respond to step-by-step prompts. Additionally, we noticed that Step-by-Step Thinking prompts also did not improve the performance of text-davinci-002, even though it is based on GPT-3.5-turbo. As there are no publicly available details about this model, we speculate, based on third-party information, that as an intermediate state model, its optimization objective might have reduced its capability to follow instructions. However, Step-by-Step prompts had a pronounced impact on GPT-3.5-turbo, which increased the correlation between humans and the model, indicating substantial progress in the model's ability to mimic human emotional understanding and thought processes.

The combination of Two-shot Chain of Thought Reasoning and Step-by-Step Thinking prompts did not lead to higher EQ scores for models like GPT-3.5-turbo, text-davinci-001, and text-davinci-003. However, it did result in increased pattern similarity. This result aligns with the official statements from OpenAI about the impact of instruct fine-tuning and RLHF techniques in making models' responses more human-like. It also suggests that these models have the potential to master patterns of emotional understanding that are similar to those used by humans.

The variance in response to different prompting techniques among models emphasizes the importance of a deeper understanding of factors such as model architecture, training data, fine-tuning process, and optimization objectives. The interplay of these factors might influence a model's receptiveness and response to different prompting techniques.

Looking forward, there is a need for further exploration into the impact of various types of prompts on LLMs during emotional intelligence tests, including the investigation of more diverse categories of prompts or hybrid prompts. In-depth studies into why certain models respond more favorably to specific prompts can also inform the development of more advanced LLMs with superior human emotional understanding capabilities. These studies could also provide valuable insights for optimizing the instructive fine-tuning process and the application of Reinforcement Learning from Human Feedback (RLHF) techniques. Ultimately, enhancing our understanding of the relationship between prompts and LLM performance in emotional intelligence tests can significantly contribute to the ongoing evolution and refinement of these models.

Table S1: Results with the direct prompts

| Based | Models | SECEU score | EQ EQ | EQ % | Pattern Similarity r | Pattern Similarity % |
|---|---|---|---|---|---|---|
| **OpenAI GPT series** | | | | | | |
| | DaVinci | | ------FAILED------ | | | |
| | Curie | | ------FAILED------ | | | |
| | Babbage | | ------FAILED------ | | | |
| | text-davinci-001 | | ------FAILED------ | | | |
| | text-davinci-002 | | ------FAILED------ | | | |
| | text-davinci-003 | 2.56 | 104 | 55% | -0.11 | 4% |
| | GPT-3.5-turbo | 2.63 | 103 | 52% | 0.04 | 17% |
| | GPT-4 | 1.89 | 117 | 89% | 0.28 | 67% |
| **LLaMA** | | | | | | |
| | LLaMA | | ------FAILED------ | | | |
| | Alpaca | 2.56 | 104 | 56% | 0.03 | 15% |
| | Vicuna | 2.50 | 105 | 59% | -0.02 | 10% |
| | Koala | 3.72 | 83 | 13% | 0.43** | 93% |
| **Flan-t5** | | | | | | |
| | Fastchat | | ------FAILED------ | | | |
| **Pythia** | | | | | | |
| | Dolly | 2.89 | 98 | 38% | 0.26 | 62% |
| | Oasst | 2.41 | 107 | 64% | 0.24 | 59% |
| **GLM** | | | | | | |
| | ChatGLM | 3.12 | 94 | 28% | 0.09 | 24% |
| **RWKV** | | | | | | |
| | RWKV-v4 | | ------FAILED------ | | | |
| **Claude** | | | | | | |
| | Claude | 2.46 | 106 | 61% | 0.11 | 28% |

Table S1 Footnote: Table S1 shows the SECEU scores, EQ scores, pattern similarity, and properties of mainstream LLMs which we evaluated in the current study under the direct prompts.
evaluated in the current study.
Failed: The LLMs cannot complete the test.
%: The percent of humans whose performance was below that of an LLM in the test.
Pattern Similarity: The degree of similarity is indexed Pearson correlation coefficient ($r$).
*: $p < 0.05$; **: $p < 0.01$.

Table S2: Results with the assistance of prompt

| Models | Prompts | SECEU score | EQ | | Pattern Similarity | |
|---|---|---|---|---|---|---|
| | | | EQ | % | r | % |
| DaVinci | | | | | | |
| | Zero-Shot | | ------FAILED------ | | | |
| | Zero-Shot + Step-by-step | | ------FAILED------ | | | |
| | Two shot COT | 3.50 | 87 | 18% | 0.41** | 91% |
| | Two shot COT + Step-by-step | 3.76 | 82 | 12% | 0.00 | 11% |
| Curie | | | | | | |
| | Zero-Shot | | ------FAILED------ | | | |
| | Zero-Shot + Step-by-step | | ------FAILED------ | | | |
| | Two shot COT | 2.82 | 100 | 42% | 0.27 | 66% |
| | Two shot COT + Step-by-step | 2.70 | 102 | 50% | 0.11 | 29% |
| Babbage | | | | | | |
| | Zero-Shot | | ------FAILED------ | | | |
| | Zero-Shot + Step-by-step | | ------FAILED------ | | | |
| | Two shot COT | 2.78 | 100 | 44% | -0.12 | 4% |
| | Two shot COT + Step-by-step | | ------FAILED------ | | | |
| GPT-3.5-turbo | | | | | | |
| | Zero-Shot | 3.12 | 94 | 28% | -0.05 | 7% |
| | Zero-Shot + Step-by-step | 2.53 | 105 | 57% | 0.02 | 14% |
| | Two shot COT | 3.08 | 95 | 30% | 0.18 | 45% |
| | Two shot COT + Step-by-step | 2.78 | 100 | 43% | 0.29 | 70% |
| text-davinci-001 | | | | | | |
| | Zero-Shot | | ------FAILED------ | | | |
| | Zero-Shot + Step-by-step | | ------FAILED------ | | | |
| | Two shot COT | 2.53 | 105 | 57% | 0.12 | 31% |
| | Two shot COT + Step-by-step | 2.40 | 107 | 64% | 0.20 | 47% |
| text-davinci-002 | | | | | | |
| | Zero-Shot | | ------FAILED------ | | | |
| | Zero-Shot + Step-by-step | | ------FAILED------ | | | |
| | Two shot COT | 3.30 | 91 | 23% | -0.04 | 8% |
| | Two shot COT + Step-by-step | 3.30 | 91 | 23% | -0.26 | 1% |
| text-davinci-003 | | | | | | |
| | Zero-Shot | | ------FAILED------ | | | |
| | Zero-Shot + Step-by-step | | ------FAILED------ | | | |
| | Two shot COT | 2.01 | 114 | 83% | 0.31* | 73% |
| | Two shot COT + Step-by-step | 2.45 | 106 | 61% | 0.33 | 79% |

Table S2 Footnote: Table S2 shows the SECEU scores, EQ scores, pattern similarity, and properties of OpenAI GPT series with the assistance of prompt.

Failed: The LLMs cannot complete the test.

%: The percent of humans whose performance was below that of an LLM in the test.

Pattern Similarity: The degree of similarity is indexed Pearson correlation coefficient ($r$).

*: $p < 0.05$; **: $p < 0.01$.

Figure S1: The distribution of individual performance (i.e., the Euclidean distance between the individual's answer and the objective standards) on each item (M±SD)

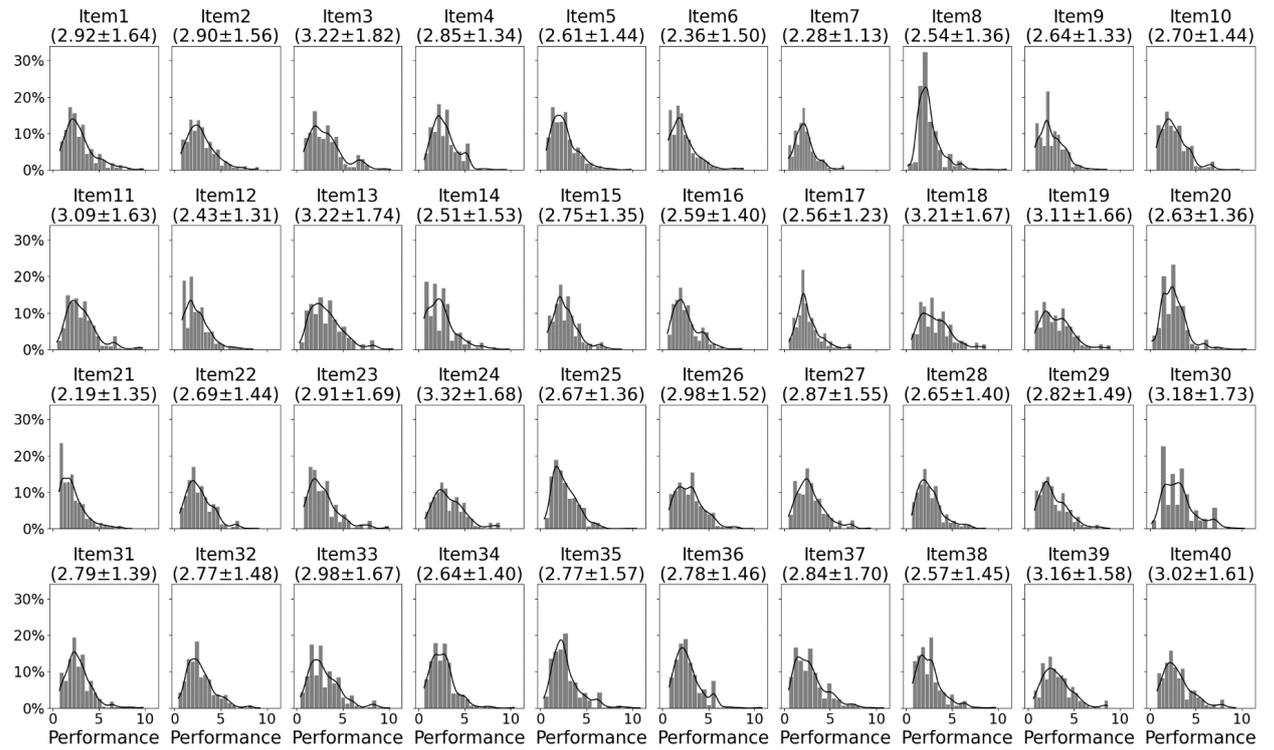